\documentclass[review]{elsarticle}

\usepackage{hyperref}
\usepackage{csquotes}
\graphicspath{ {images/} }

\usepackage{adjustbox}

\journal{Journal of \LaTeX\ Templates}

\begin{document}

\begin{frontmatter}

%\title{Elsevier \LaTeX\ template\tnoteref{mytitlenote}}
%\tnotetext[mytitlenote]{Fully documented templates are available in the elsarticle package on %\href{http://www.ctan.org/tex-archive/macros/latex/contrib/elsarticle}{CTAN}.}
\title{DeepRare : Generic Unsupervised \\
Visual Attention Models}

%% Group authors per affiliation:
%\author{Elsevier\fnref{myfootnote}}
%\address{Radarweg 29, Amsterdam}
%\fntext[myfootnote]{Since 1880.}

%% or include affiliations in footnotes:
%\author[mymainaddress,mysecondaryaddress]{Elsevier Inc}
%\ead[url]{www.elsevier.com}

%\author[mysecondaryaddress]{Global Customer Service\corref{mycorrespondingauthor}}
%\cortext[mycorrespondingauthor]{Corresponding author}
%\ead{support@elsevier.com}

%\address[mymainaddress]{1600 John F Kennedy Boulevard, Philadelphia}
%\address[mysecondaryaddress]{360 Park Avenue South, New York}

\author[mymainaddress,mysecondaryaddress]{Phutphalla Kong}

\author[mysecondaryaddress]{Matei Mancas\corref{mycorrespondingauthor}}
\cortext[mycorrespondingauthor]{Corresponding author}
\ead{matei.mancas@umons.ac.be}

\author[mysecondaryaddress]{Bernard Gosselin}

\author[mymainaddress]{Kimtho Po}

\address[mymainaddress]{Institute of Technology of Cambodia}
\address[mysecondaryaddress]{University of Mons, ISIA Lab}

\begin{abstract}
Human visual system is modeled in engineering field providing feature-engineered methods which detect contrasted/surprising/unusual data into images. This data is \enquote{interesting} for humans and leads to numerous applications. Deep learning (DNNs) drastically improved the algorithms efficiency on the main benchmark datasets. However, DNN-based models are counter-intuitive: surprising or unusual data is by definition difficult to learn because of its low occurrence probability. In reality, DNN-based models mainly learn top-down features such as faces, text, people, or animals which usually attract human attention, but they have low efficiency in extracting surprising or unusual data in the images. In this paper, we propose a new visual attention model called DeepRare2021 (\textbf{DR21}) which uses the power of DNNs feature extraction and the genericity of feature-engineered algorithms. This algorithm is an evolution of a previous version called DeepRare2019 (\textbf{DR19}) based on a common framework. \textbf{DR21} 1) does not need any training and uses the default ImageNet training, 2) is fast even on CPU, 3) is tested on four very different eye-tracking datasets showing that the \textbf{DR21} is generic and is always in the within the top models on all datasets and metrics while no other model exhibits such a regularity and genericity. Finally \textbf{DR21} 4) is tested with several network architectures such as VGG16 (V16), VGG19 (V19) and MobileNetV2 (MN2) and 5) it provides explanation and transparency on which parts of the image are the most surprising at different levels despite the use of a DNN-based feature extractor. DeepRare2021 code can be found at \url{https://github.com/numediart/VisualAttention-RareFamily}.
\end{abstract}


\end{frontmatter}

%\linenumbers

\section{Visual attention: deep learning trouble}
\label{sec:intro}
Human visual system handles a huge quantity of incoming visual information and it cannot carry out multiple complex tasks in the same time on the whole visual field. This bottleneck \cite{donald1958} implies that it has an exceptional ability of sampling the surrounding world and pay attention to objects of interest. In computer vision, visual attention is mainly modeled through the so-called saliency maps. The modeling of visual attention has numerous applications such as object detection, image segmentation, image/video compression, robotics, image re-targeting, visual marketing and so on \cite{mancas2016}. Visual attention is considered to be a mix of bottom-up and top-down information. Bottom-up information is based on low-level features such as luminance, chrominance, or texture. Top-down information is more related to knowledge people already have about their tasks or objects they see such as faces, text, persons, or animals.

\iffalse
    In computer vision, visual attention is modeled through the so-called saliency maps. The modeling of visual attention has numerous applications such as object detection \cite{butko2009}, \cite{Ehinger2009}, image segmentation \cite{mishra2009}, \cite{maki2000}, image/video compression \cite{itti2004}, \cite{guo2010}, image re-targeting \cite{mar2009}, \cite{suh2003}, \cite{mancas2016}, and so on. 
\fi

\iffalse
    \subsection{The age of feature-engineered saliency}
\fi

Since the early 2000, numerous models of visual attention based on image features were provided. In this paper, they will be referred as \enquote{classical models}. While they can be very different in their implementation, most of them have the same main philosophy: search for contrasted, rare, abnormal or surprising features within a given context. Among those models one may find seminal work of \cite{itti2000} or \cite{rose1999}, but also more recent work based on information processing such as AIM \cite{aim}. Finally, some models became a reference for classical models such as GBVS \cite{gbvs}, RARE \cite{rare2012}, BMS \cite{bms2013} or AWS \cite{aws}. 

\iffalse
    \subsection{The rise of deep learning}
\fi

With the arrival of the deep learning wave, most researchers have focused on Deep Neural Networks saliency which will be referred as \enquote{DNN-based} in this paper. DNN-based models triggered a revolution in terms of results on the main benchmark datasets such as MIT benchmark \cite{mit-saliency-benchmark} where DNN-based saliency models definitely outperformed classical models. The DNN-based models have been already used in several applications such as image and video processing, medical signal processing or big data analysis \cite{sun15}, \cite{zhao15}, \cite{qin15}, \cite{han14}, \cite{sun16}. Some of the DNN-based models became new references such as SALICON \cite{salicon2015}, MLNet \cite{mlnet2016} or SAM-ResNet \cite{sam2018}.

\iffalse
    \subsection{Trouble into the deep learning}
\fi

However, recently DNN-based models have been criticized for some drawbacks. 
First, they underestimate the importance of bottom-up attention \cite{kum17} which indicates that they were mostly trained to detect the attractive top-down objects rather than detect saliency itself. In \cite{kong19} the authors found that if saliency models very precisely detect top-down features, they neglect a lot of bottom-up information which is surprising and rare, thus by definition difficult to learn. This shows that saliency cannot be learnt but instead objects \cite{kong18} which are often attended by human gaze (such as faces, text, bodies, etc.) are learnt and by the way, they are enough to provide good results on the main benchmarks. 

A second drawback of the DNN-based models is that in addition to not take into account low-level features surprise level, DNN-based models are not generic enough to adapt to new images which are different enough from the training dataset. Indeed, recently, \cite{Kotseruba2019} introduced two novel datasets, one based on psycho-physical patterns (P\textsuperscript{3}) and one based on natural odd-one-out (O\textsuperscript{3}) stimuli. They showed that while DNN-based models are good in MIT dataset on natural images, their results drastically drop on P\textsuperscript{3} and O\textsuperscript{3}. 

A third drawback of the DNN-based models is linked to DNNs themselves which are black boxes. When a models fails to predict saliency, there is no way to understand why this prediction failed. 

In parallel to DNN-based models, DeepFeat \cite{deepFeat} or SCAFI \cite{scafi} deal with models where pre-trained deep features are used. Those models will be called \enquote{deep-features models} in this paper. However, they are not yet comparable to DNN-based models for general images datasets such as the MIT benchmark. Based on the new datasets in \cite{Kotseruba2019}, DeepRare2019 \cite{kong20} provides a new deep-feature saliency model by mixing deep features and the philosophy of an existing classical model \cite{rare2012}. This model is efficient on all the datasets, with no need for any training and efficient in terms of computation even on CPU. 

In this paper we build on DeepRare2019 to improve it in several ways : 1) different DNN architectures are used and compared (VGG16, VGG19 and MobileNetV2) on more datsets, 2) a threshold on the feature rarity is introduced which let us understand which parts of the image are the most surprising at different levels providing transparency to the model, 3) the best combination of thresholds and an improved post-processing which lead to results which are much better than for DeepRare2019. This new model is called DeepRare2021 (\textbf{DR21}) and shows that the DeepRare framework is modular and can easily evolve.    

\iffalse
    Finally, we can add the usual DNN black-box drawback which is the impossibility to explain the results. 
    \subsection{Paper plan}
\fi

In a section \ref{sec:DeepRare2019} \textbf{DR21} is described and the threshold on feature rarity is used to show how the DNN features rarity can become explainable. In section \ref{sec:valid} this model is tested on the datasets proposed in \cite{Kotseruba2019} but also on an additional dataset. We finally discuss and conclude on the pertinence of the come back of the feature engineering models. 

\section{DeepRare2021 model: digging into rare deep features}
\label{sec:DeepRare2019}
\iffalse
    As previously stated, classical visual attention models are generic, they generally allow to understand the results obtained are they are good in low-level feature attention. On the other side, DNN-based models provide results which cannot be explained and are sometimes tuned for specific types of datasets, however they are good in higher level and especially top-down information which is most of the time very important in terms of eye-tracking.

!!!!!! ADD thresholds and empirical choice and 3 different networks to be tested.  !!!!!!
\fi

In this paper we extend the approach in \cite{kong20} where a framework called DeepRare is proposed which mixes the simplicity of the idea of rarity computation to find the most salient features with the advantages of deep features extraction. Indeed, rare features attract human attention as they are surprising compared to the other features within the image. This combination has the advantage to be fast (less than 1 second per image on CPU with a VGG16 feature extractor) and easy to adapt to any default DNN architectures (VGG19, ResNet, etc.). Here, we extend \textbf{DR19} adding the possibility to have thresholds on the rarity maps and also the possibility to use several DNN architectures. This additional work leads to the DeepRare2021 (\textbf{DR21}) representation of the image where the features are selected based on their rarity before combining them. In the following sections we describe \textbf{DR21} and its feature visualisation.   

\subsection{DeepRare fremawork}

\begin{figure}[!ht]
\centering
\includegraphics[width=3.4in]{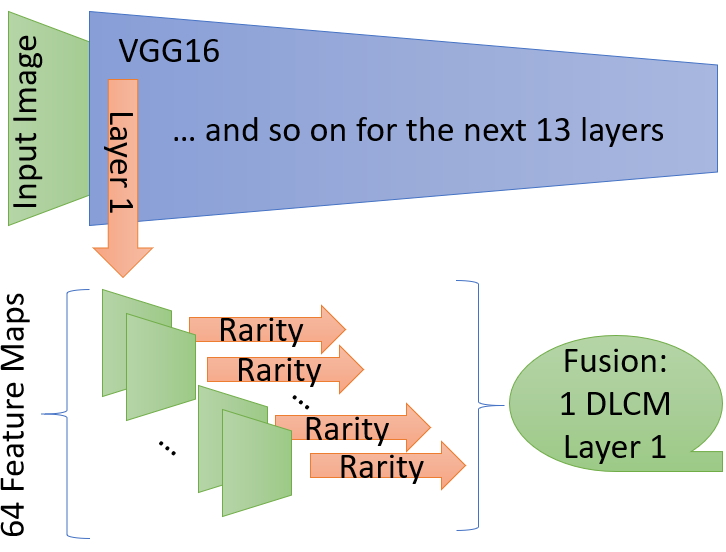}
\caption{Processing for Layer 1. This processing is iterated for all interesting layers from a CNN encoder network. In this example 13 layers are in total chosen for a VGG16 encoder.}
\label{fig:DR}
\end{figure}

Figure \ref{fig:DR} summarizes the DeepRare architecture. From an input image, features are extracted based on a CNN encoder (such as a VGG16). This network will extract features needed to solve its training task. Here, we use the image classification task on the ImageNet dataset \cite{deng2009imagenet} which is a dataset made of very diverse images and more than 1000 classes of objects. These weights are available by default in Keras \cite{chollet2015keras} or other development frameworks. Once features for some selected layers are extracted, their rarity is computed. The next step let us select the most rare features and to represent them easily. At the end the selected features are fused and post-processed in a final saliency map. All those steps are described in the following sections. 

\iffalse
A convolutional network is a great tool for feature extraction. When trained on a general dataset such as ImageNET, the network will extract a complete set of features that one finds in images at several scales (from very low-level in the first layers to very high level in the last ones). We decide here to use a VGG16 architecture with its default training on ImageNET dataset as a feature extractor, but any other architecture could be used as well. Our implementation is based on Keras framework to extract the convolutional layers and feature maps within those layers. We do not use (1) the pooling layers (as they are redundant with the previous convolutional layer) and (2) the final fully connected classification layers. An example for layer 1 is illustrated in Figure \ref{fig:DR}.

In a VGG16, the convolutional layers are gathered within 5 groups separated by the pooling layers : 1) the first low-level features in layers 1 and 2, then 2) second set of low-level features from layers 4 and 5, after that 3) the first middle-level layers 7, 8 and 9 and 4) the second middle-level layers 11, 12 and 13 and finally 5) the high-level features from layers 15, 16 and 17.

\fi

\subsection{CNN architectures and layers taken into account}
\label{subsec:takeacc}
While in \textbf{DR19}, the algorithm is applied only to a VGG16 architecture, \textbf{DR21} can be applied to various convolutional architectures. In this paper we apply it to a VGG16, VGG19, and MobileNetV2 architectures. While VGG19 is a variant of the VGG16 architecture, MobileNetV2 is very different and it has the advantage to be light in terms of weight and computation which makes it usable on embedded devices such as smartphones, etc. 

The rarity is not computed on all the layers to avoid adding unnecessary information. 
For VGG16, we do not use (1) the pooling layers (as they are redundant with the previous convolutional layer) and (2) the final fully connected classification layers. In a VGG16, the convolutional layers are gathered within 5 groups separated by the pooling layers : 1) the first low-level features in layers 1 and 2, then 2) second set of low-level features from layers 4 and 5, after that 3) the first middle-level layers 7, 8 and 9 and 4) the second middle-level layers 11, 12, and 13 and finally 5) the high-level features from layers 15, 16, and 17. For VGG19, the same approach was taken into account. We take layer 1 and 2 for the first low-level features; layer 4 and 5 for the second low-level features; layer 7, 8, 9, and 10 for the first middle-level features; layer 12, 13, 14, and 15 for the second middle-level features; and layer 17, 18, 19, and 20 for the high-level features. For MobileNetV2, we use the same approach as VGG16 and VGG19. However, the architecture is much more complex. We take layer 16 and 18 for the first low-level features; layer 24 and 32 for the second low-level features; layer 41, 50, 59, and 67 for the first middle-level features; layer 76, 85, 94, and 102 for the second middle-level features; and layer 111, 120, 137, and 146 for the high-level features.

\subsection{Rarity of deep features and top-down information}
Once, the layers taken into account in the algorithm are selected for the given CNN architecture, it is necessary to compute the feature maps rarity within those layers. Figure \ref{fig:DR} shows that, on each feature map within a selected layer we compute the data rarity. For that, as in \textbf{DR19}, we use the main idea from \cite{rare2012} without the multi-resolution part which is naturally achieved by the convolutional network architecture. A very simple rarity function \textit{R} based on the histogram of each feature map sampled on a few bins (11 in the current implementation) is used as in equation \ref{eqn1}.
\begin{eqnarray}
\label{eqn1}
   R(i) = -log(p(i))
\end{eqnarray}
where \textit{p(i)} is the occurrence probability for the pixels of bin \textit{i}. 
Once the rarity histogram \textit{R} is computed, the resulting rarity image is reconstructed by backprojection. This operation uses the histogram of a feature (here the rarity of a feature) to find this feature in an image projecting each histogram value on the corresponding pixel in an image. This image will highlight pixels in the feature map which are rare compared to the other pixels in the feature map. Based on \cite{rare2012}, rare pixels are the ones which might attract human attention. 

The advantage of this approach is that it is very fast to compute and this is important as it needs to be applied to numerous feature maps.

\subsection{Digging into Rare Deep Features}
\label{sec:feature}

Once we decided the layers which will be taken into account into the model and we computed their rarity, we can go further and select the most rare features in the feature maps.  
In that aim we decided to apply a threshold on the computed rarity maps. This threshold is applied directly on the rarity of each feature map and varies from 0 (no threshold) to 0.9 (only keeping the 10\% most rare features) by steps of 0.1. A binary threshold is first obtained and used as a mask on the feature map to keep only the values within this mask while the rest are set to 0.  

\begin{figure}[!ht]
\centering
\includegraphics[width=4.7in]{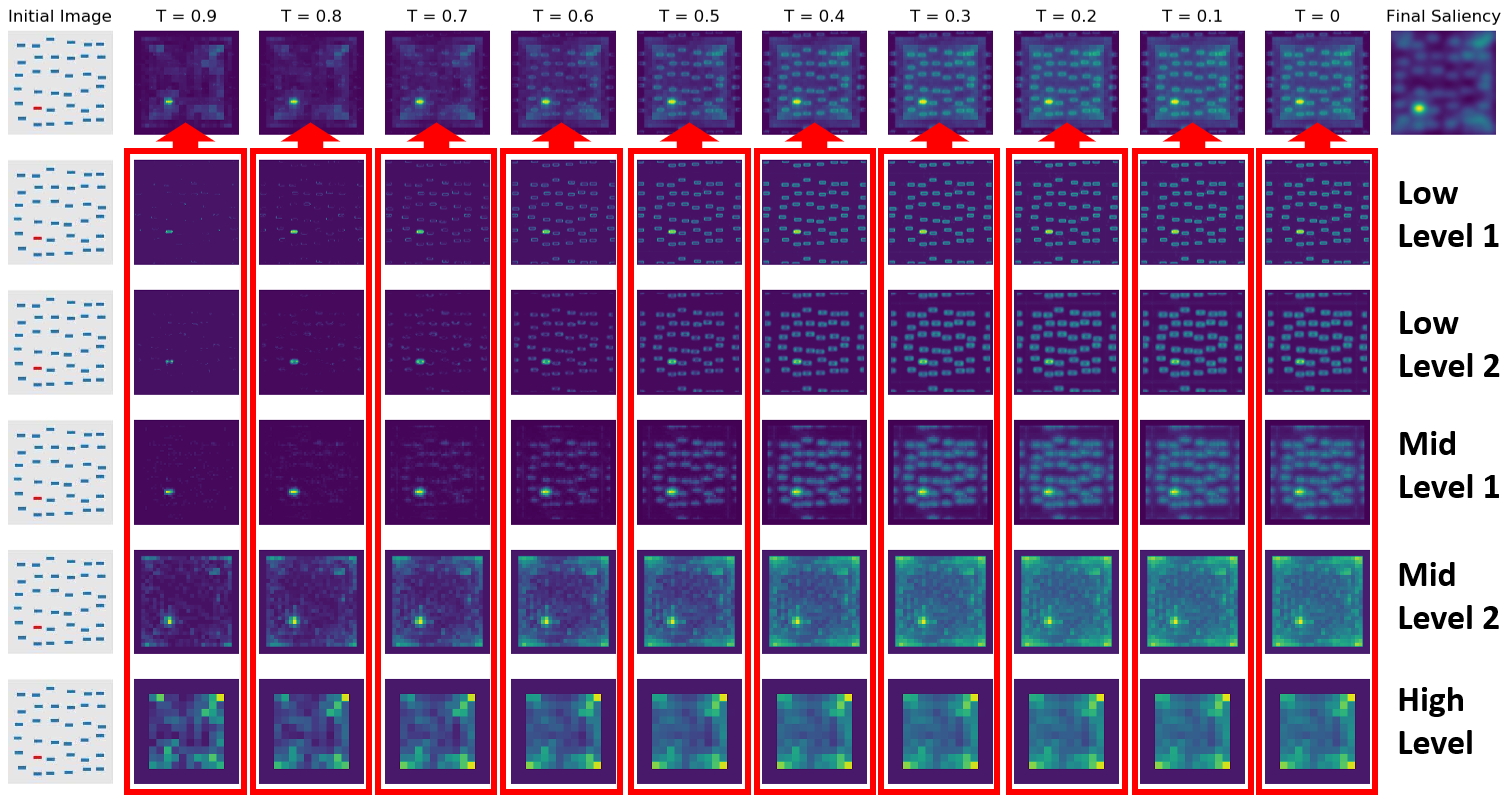}
\caption{Detailed maps of different levels (from Low Level 1 to High Level) and different thresholds on feature rarity (from 0.9 to no threshold) within the VGG16 architecture.}
\label{fig:feat1}
\end{figure}

\begin{figure}[!ht]
\centering
\includegraphics[width=4.7in]{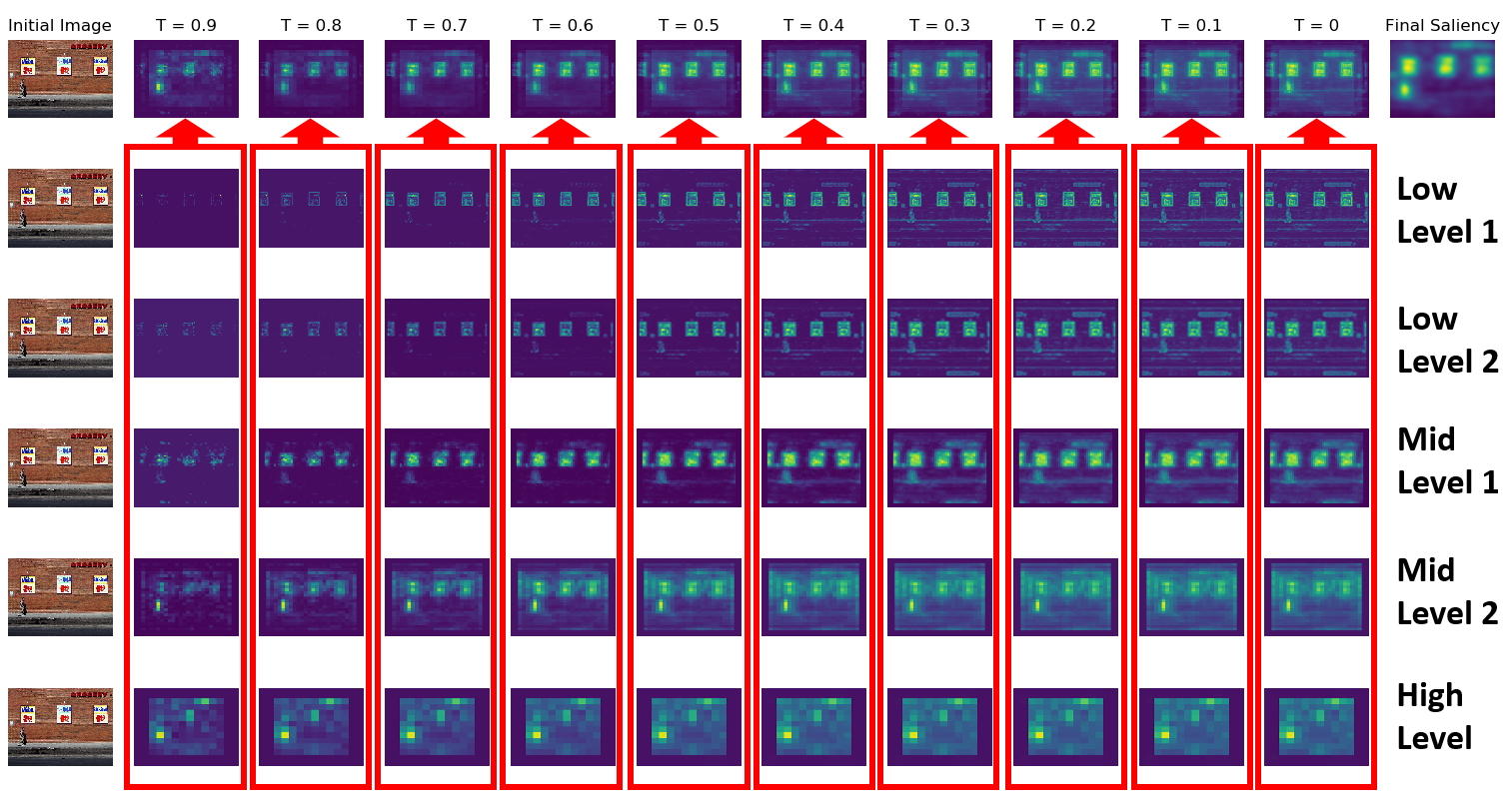}
\caption{Detailed maps of different levels (from Low Level 1 to High Level) and different thresholds on feature rarity (from 0.9 to no threshold) within the VGG16 architecture.}
\label{fig:feat2}
\end{figure}

\begin{figure}[!ht]
\centering
\includegraphics[width=4.7in]{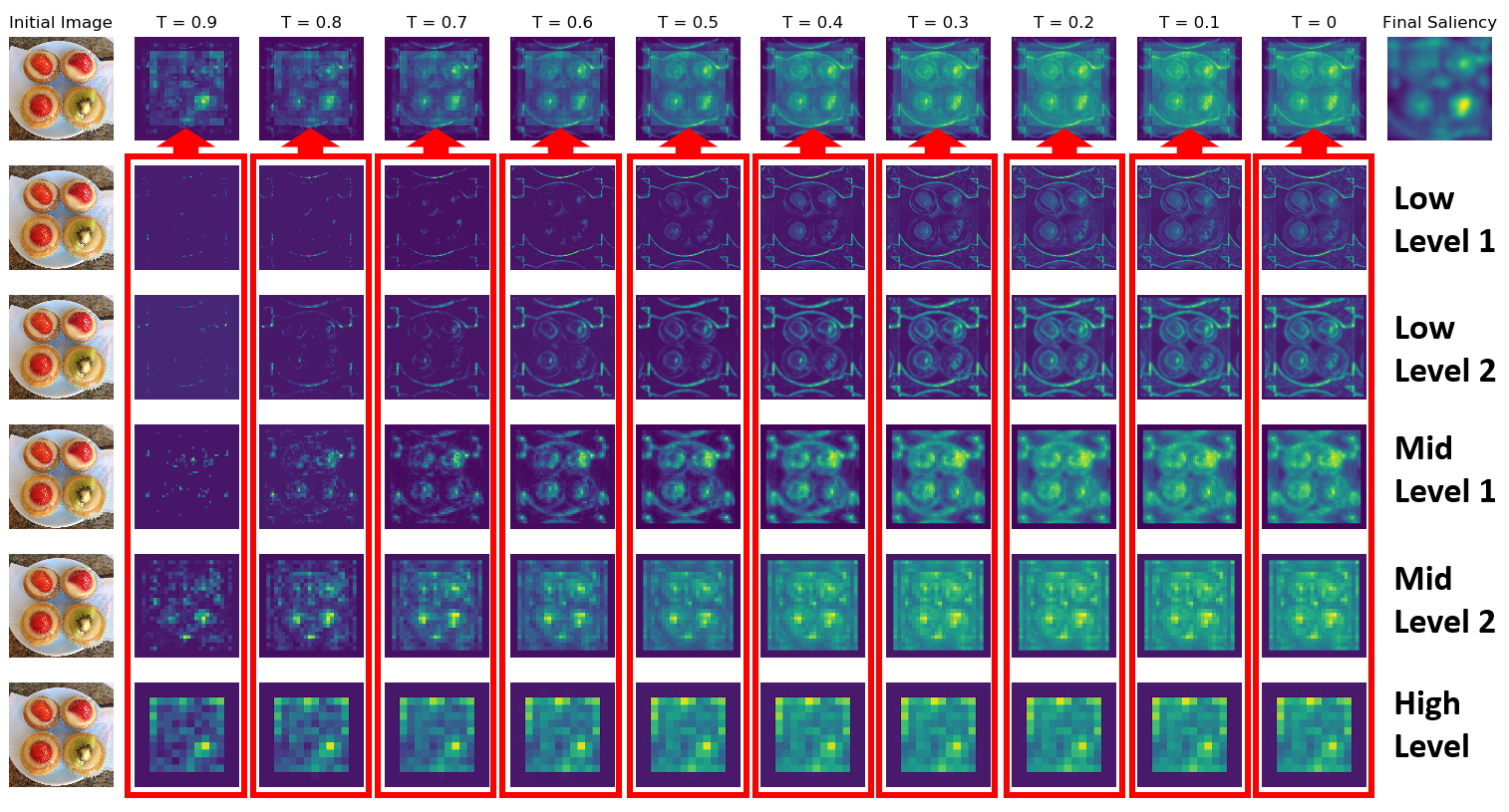}
\caption{Detailed maps of different levels (from Low Level 1 to High Level) and different thresholds on feature rarity (from 0.9 to no threshold) within the VGG16 architecture.}
\label{fig:feat3}
\end{figure}

In this section we inspect the rare deep features at different scales to understand what this rarity thresholds physically mean. One advantage of \textbf{DR21} is that it is possible to investigate at which scale and where the feature rarity is important and thus let us understand how the attention mechanism works and how the image structures are taken into account. In this section, figures \ref{fig:feat1}, \ref{fig:feat2}, \ref{fig:feat3} are computed with a VGG16 architecture and the 5 groups discussed in section \ref{subsec:takeacc}. 

In Fig. \ref{fig:feat1} we inspect a simple image with an obvious low-level focus of attention. The initial image (on the left) represents several horizontal blue bars while only one is in red. This red bar is an obvious point of attention based on a low-level feature : the color. 

From this image, there are 10 columns with different thresholds from T = 0.9 which only keep the 10\% rarest features to T = 0 where no threshold was applied to the rarity feature maps. Lines 2 to 6 represent the features for different levels (5 levels when using a VGG16 architecture) which are already a fusion of the selected layers rarity maps (for the fusion, see next section). The final fusion of the 5 levels can be found on the first line. The post-processing saliency map (see section \ref{sec:smpp}) can be found on top-right of the image.  

For the higher threshold (T=0.9) the abnormal region is detected on all levels except the higher level where the edge effects are too important (and can be seen in the corners even when the edges of the image are set to 0). For the low levels (such as level 1 to level 3) only the red pattern appears and the model is very precise and selective on the rare object. When going towards the right with lower thresholds, little by little, the other blue patterns also appear while the red one is still the most highlighted but the distractors around are visible.    

In Fig. \ref{fig:feat2} one can see the result for a situation where mid-level (big letters) and high-level features (such as text and people) are the rare features (see initial image on the left). This image has a less obvious attention focus as the one in Fig. \ref{fig:feat1}. 

For the higher threshold (T=0.9) the abnormal regions are split between mid levels and the higher level. While at the low-levels very few information passes the threshold, for the higher levels text and the person are well highlighted. For the last level the bigger text and the person are more highlighted than small text. At smaller thresholds the low levels highlight mostly the posters on the wall based on their colors but not enough the person and the large letters on top-right.

In Fig. \ref{fig:feat3} one can see the result for a situation where high-level features (big cake shape and color) are the rare features. For the higher threshold (T=0.9) the abnormal regions are only detected in the higher level (mid level 2 and especially high level). On all the other levels no interesting feature is highlighted. For small thresholds, for low level 1 and 2 and mid level 1 only edges and object areas are highlighted but the model fails in detecting the different cake. We see that here, the low level feature never detect the abnormal cake whatever the threshold is.    

Overall in figures \ref{fig:feat1}, \ref{fig:feat2} and \ref{fig:feat3} mid level 2 and high level provide always better results with a high threshold such as T = 0.9, while lower level feature work better on this high threshold only in specific kind of images with obvious abnormal patterns dues to low-level features. We already understand from here that several thresholds need to be combined to provide better final results. 

In \cite{kong19} the authors showed that top-down information for high-level features such as text, people, animals or transportation had a huge impact on visual attention through the mix of those features with a simple rarity bottom-up approach. But those rarity-based features were only low-level features. 
In the current paper we use both mid-level and high-level features however we do not add top-down information (except for a weak face detector only added when the VGG16 architecture is used).
In the following section we show how the thresholded rarity feature maps from the chosen layers are fused together.

\subsection{Data fusion}
\label{datafusion}
Once the rarity of all feature maps is computed, the results need to be fused together. We use a classical map fusion from \cite{itti2004} where the fusion weights depend on the squared difference between the max and the mean of each map. This is applied to all feature maps within each layer leading to the deep layer conspicuity maps (DLCM), one for each convolutional layer (see Figure \ref{fig:DR} for first layer). This approach is efficient and simple which is good as it needs to be applied an important amount of time. 

In a second stage, the same fusion method is applied for each of the layer groups arriving to 5 deep groups conspicuity maps (DGCM). This fusion is made in a way to give more importance to higher level layers.

Finally, the 5 DGCM are summed up. In the case a VGG16 architecture is used, a top-down face map can be added based on feature map \#105 from layer 15 which is known to detect faces which are large enough \cite{scafi}. 

We show here different configurations of thresholds on the layers and check the results for the VGG16 architecture (tables \ref{tab:dfo} and \ref{tab:dfm}). The accuracy is here computed by using the correlation metric (CC) between the final saliency map and the real people gaze obtained by using eye-tracking. 

\begin{table}[!t]
\begin{center}
\caption{The OSIE dataset. It applies both face and without face features on VGG16.}
\label{tab:dfo}
\begin{tabular}{|c|c|c|}
\hline
VGG16 & With face & Without face\tabularnewline
\hline 
Thresholds & CC & CC\tabularnewline
\hline 
0 & 0.55 & 0.53\tabularnewline
\hline 
0.9 & 0.56 & 0.55\tabularnewline
\hline 
(0+0.9)/2 & 0.57 & 0.56\tabularnewline
\hline 
(0.4+0.9)/2 & 0.57 & 0.56\tabularnewline
\hline 
\end{tabular}
\end{center}
\end{table}

\begin{table}[!t]
\begin{center}
\caption{The MIT1003 dataset. It applies both face and without face features on VGG16.}
\label{tab:dfm}
\begin{tabular}{|c|c|c|}
\hline 
VGG16 & With face & Without face\tabularnewline
\hline 
Thresholds & CC & CC\tabularnewline
\hline 
0 & 0.47 & 0.46\tabularnewline
\hline 
0.9 & 0.45 & 0.43\tabularnewline
\hline 
(0+0.9)/2 & 0.48 & 0.47\tabularnewline
\hline 
(0.4+0.9)/2 & 0.47 & 0.45\tabularnewline
\hline 
\end{tabular}
\end{center}
\end{table}

We observe that on two different validation datasets with natural images (OSIE and MIT1003) the use of the face improves the results. On OSIE dataset, the use of the higher threshold (0.9) or no threshold (0) has different effects producing better results on the thresholded rarity layers on OSIE (table \ref{tab:dfo}) and less good resuls on MIT1003 (table \ref{tab:dfm}). However, the combination of the thresholds 0 and 0.9 is better in both cases while the combination between 0 and 0.4 is a little less good on images from MIT1003. These tests show that it always works better to mix the 0 threshold which shows all the data classified by order of rarity and the 0.9 which is the higher threshold which only lets the most rare regions pass. At the end we have the best mix which is to take into account all the rare data (threshold 0) and reinforce the areas with very rare data (threshold 0.9).    

\subsection{Saliency map post processing}
\label{sec:smpp}
Once maps were fused, it is well known \cite{Judd_2012} that a post-processing of the saliency maps can improve the final results depending on the validation metrics. Indeed, the eye-tracking data which is used for validation leads to rather fuzzy eye-tracking saliency maps, thus the correlation with fuzzy predicted saliency maps will be better. Here we used a gaussian low-pass smoothing filtering approach to optimize the final saliency map with the same parameters as in \cite{kong19}.

In addition of smoothing the data we tested the fact of squaring the data after the smoothing. Tables \ref{tab:smo} and \ref{tab:smm} show the results for the chosen configuration in section \ref{datafusion} which is the mix of threshold 0 and 0.9 1) not filtered, 2) using the filtering technique from \cite{kong20} and 3) squared after the filtering technique. We can see that in all cases the filter followed by the square provides the best results. When trying to put the image at power 3 or more, results are less good so we decided to keep as the final post processing scheme the filtering from \cite{kong20} followed by the squared map. 

\begin{table}[!t]
\begin{center}
\caption{The OSIE dataset. It tests on threshold 0 and 0.9 by considering on without filtering, filtering, and filtering in power 2.}
\label{tab:smo}
\begin{tabular}{|c|c|c|}
\hline 
VGG16 & With face & Without face\tabularnewline
\hline 
(0+0.9)/2 & CC & CC\tabularnewline
\hline 
No filtered & 0.54 & 0.53\tabularnewline
\hline 
Filtered & 0.57 & 0.56\tabularnewline
\hline 
Filtered + squared & 0.59 & 0.58\tabularnewline
\hline 
\end{tabular}
\end{center}
\end{table}

\begin{table}[!t]
\begin{center}
\caption{The MIT1003 dataset. It tests on threshold 0 and 0.9 by considering on without filtering, filtering, and filtering in power 2.}
\label{tab:smm}
\begin{tabular}{|c|c|c|}
\hline 
VGG16 & With face & Without face\tabularnewline
\hline 
(0+0.9)/2 & CC & CC\tabularnewline
\hline 
No filtered & 0.43 & 0.42\tabularnewline
\hline 
Filtered & 0.48 & 0.47\tabularnewline
\hline 
Filtered + squared & 0.51 & 0.50\tabularnewline
\hline 
\end{tabular}
\end{center}
\end{table}

\section{Experiments and results}
\label{sec:valid}
\iffalse
    To validate our results, we compare DR to several classical and DNN-based models both in a qualitative and quantitative way on three datasets also used in \cite{Kotseruba2019} which are P3, O3 and MIT1003. In addition we also compare our results with the DeepFeat \cite{deepFeat} model on MIT1003 dataset.

    \subsection{Data and Metrics for Validation}
    \label{sec:db}

\fi
We use 4 datasets namely OSIE \cite{OSIE}, MIT1003 \cite{Judd2009}, P\textsuperscript{3}, and O\textsuperscript{3} datasets \cite{Kotseruba2019} to validate our results. The OSIE dataset contains information at three levels: pixel-level image attributes, object-level attributes, and semantic-level attributes. The MIT1003 dataset contains general-purpose real-life images but has no specific categories or attributes. The P\textsuperscript{3} dataset evaluates the ability of saliency algorithms to find singleton targets which focuses on color, orientation, and size (without center bias). The O\textsuperscript{3} dataset depicts a scene with multiple objects similar to each other in appearance (distractors) and a singleton (target) which focuses on color, shape, and size (with center bias). We decided to use these 4 very different datasets to check how saliency models behave when facing images in different contexts. 

Concerning metrics, we use measures from \cite{Kotseruba2019}. The \enquote{number of fixations} (\# fix.) is defined as the path formed by the saliency maximum followed by the other maxima of the saliency map before reaching the target. The global saliency index (GSI) measures how well the target mean saliency is distinguished from the distractors. The maximum saliency ratio (MSR) focuses on maximum saliency of the target versus the distractors \cite{Wloka2016} and the same for the background versus target (MSR\textsubscript{b} and MSR\textsubscript{t}). We also use standard eye-tracking evaluation metrics from MIT benchmark \cite{mit-saliency-benchmark} such as Correlation Coefficient (CC), Kullback–Leibler divergence (KL), Area Under the ROC Curve from Judd (AUCJ), Area Under the ROC Curve from Borji (AUCB), Normalized Scan-path Saliency (NSS), and Similarity (SIM).

\subsection{Qualitative validation on the different datasets}
We compare our model to other models on P\textsuperscript{3} and O\textsuperscript{3} datasets. According to \cite{Kotseruba2019}, they observe that most classical models  perform better on P\textsuperscript{3} than DNN-based models. In contrast, DNN-based models perform better on O\textsuperscript{3}. 

\begin{figure}[!ht]
\centering
\includegraphics[width=4.5in]{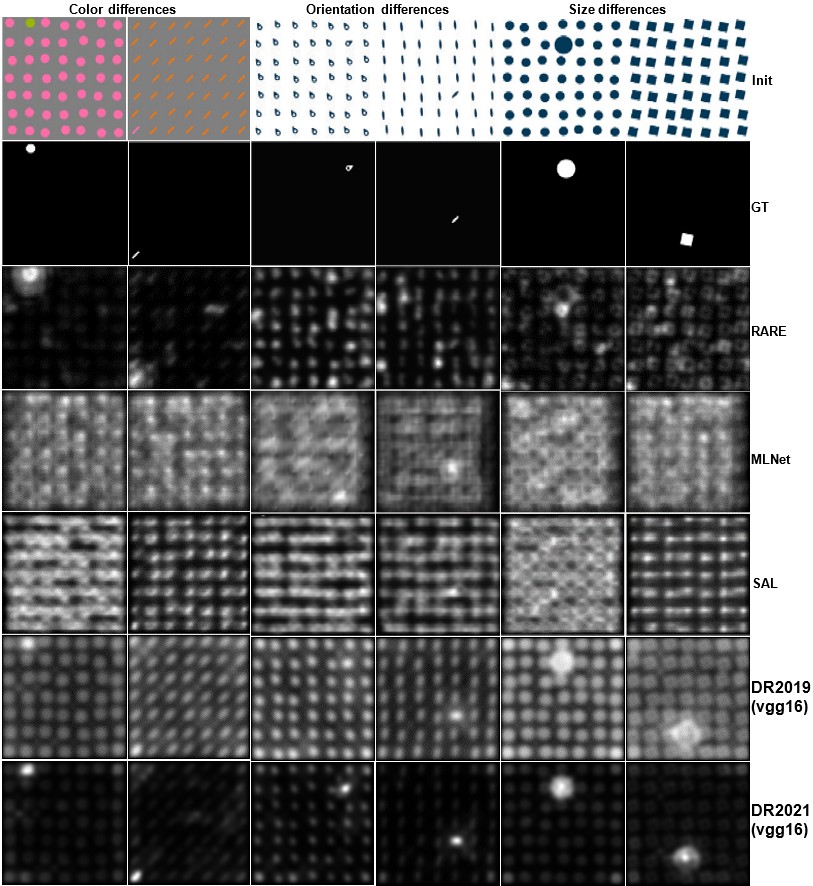}
\caption{Selected samples P\textsuperscript{3} dataset. From left to right : target difference in color, orientation, and size. From top to down : initial, ground truth, RARE2012, MLNET, SALICON, DR2019, DR2021.}
\label{fig:fig3sup}
\end{figure}

Figure \ref{fig:fig3sup} shows six samples from P\textsuperscript{3} dataset which exhibit color, orientation, and size differences of the target. While distractors are still visible on \textbf{DR19} saliency map, the targets are always correctly highlighted compared to RARE2012 which works well mainly for colors and two DNN-based models (MLNet and SALICON) which only work on one sample. \textbf{DR21} also spots all the targets but in addition, it highly decreases the distractors influence making the results very close to the ones in line 2 (ground truth).

\begin{figure}[!ht]
\centering
\includegraphics[width=4.5in]{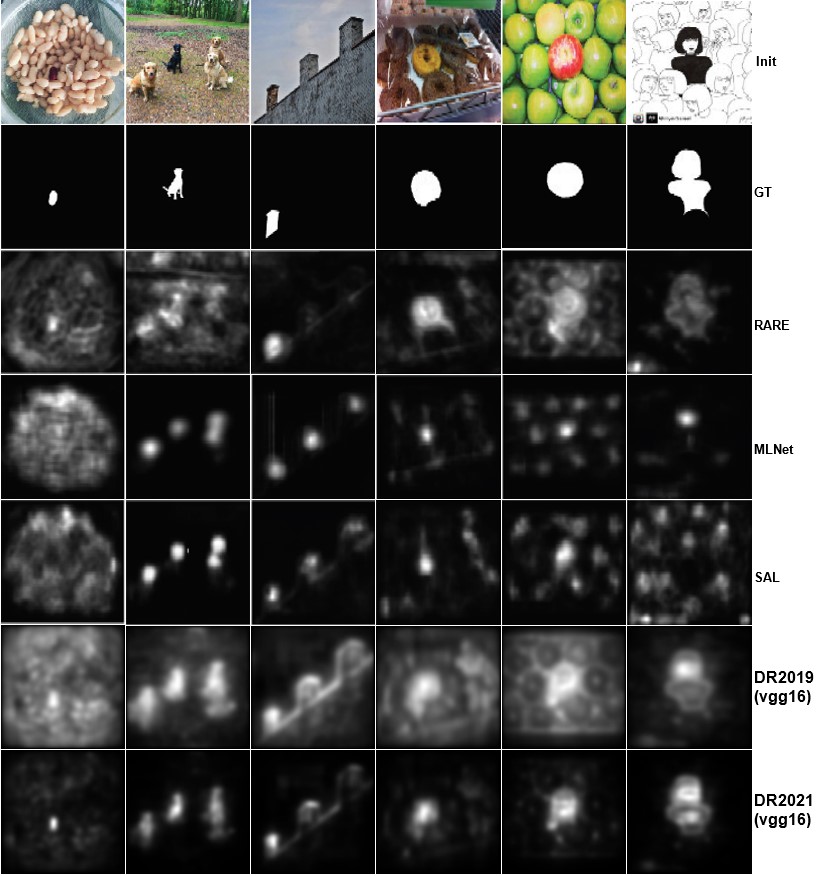}
\caption{Selected samples O\textsuperscript{3} dataset. From top to down : initial, ground truth, RARE2012, MLNET, SALICON, DR2019, DR2021.}
\label{fig:fig4sup2}
\end{figure}

Figure \ref{fig:fig4sup2} shows images from O\textsuperscript{3} dataset for different target categories (easy or difficult). Again, \textbf{DR19} highlights the target better than the DNN-based models. \textbf{DR19} seems equivalent in average with RARE. \textbf{DR21} shows again a much more precise detection eliminating distractors and background information. From a qualitative point of view, on the image in figure \ref{fig:fig4sup2}, \textbf{DR21} is the closest to the second line images (ground truth). 

\begin{figure}[!ht]
\centering
\includegraphics[width=4.5in]{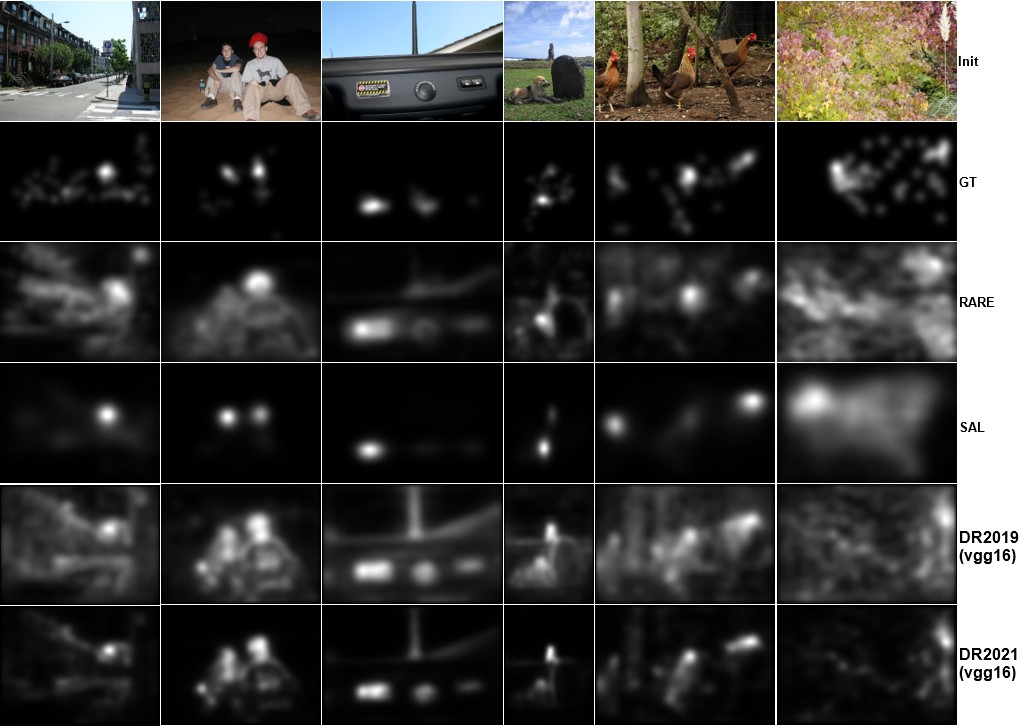}
\caption{Selected samples MIT1003 dataset. From top to down : initial, ground truth, RARE2012, SALICON, DR2019, DR2021.}
\label{fig:figMIT}
\end{figure}

Figure \ref{fig:figMIT} shows images from MIT1003 dataset. \textbf{DR19} always finds the ground truth (GT) focus regions (except for the right image where one GT focus is just in the middle probably due to the centered bias) but it also has details around those focus areas which might decrease its scores on MIT1003. \textbf{DR21} is more precise but still keeping the same focus areas. Compared to ground truth (line 2) the focus areas are the same but probably less focused as other DNN-based models which might affect its scores even if those scores should be higher than \textbf{DR19}. 

\begin{figure}[!ht]
\centering
\includegraphics[width=4.5in]{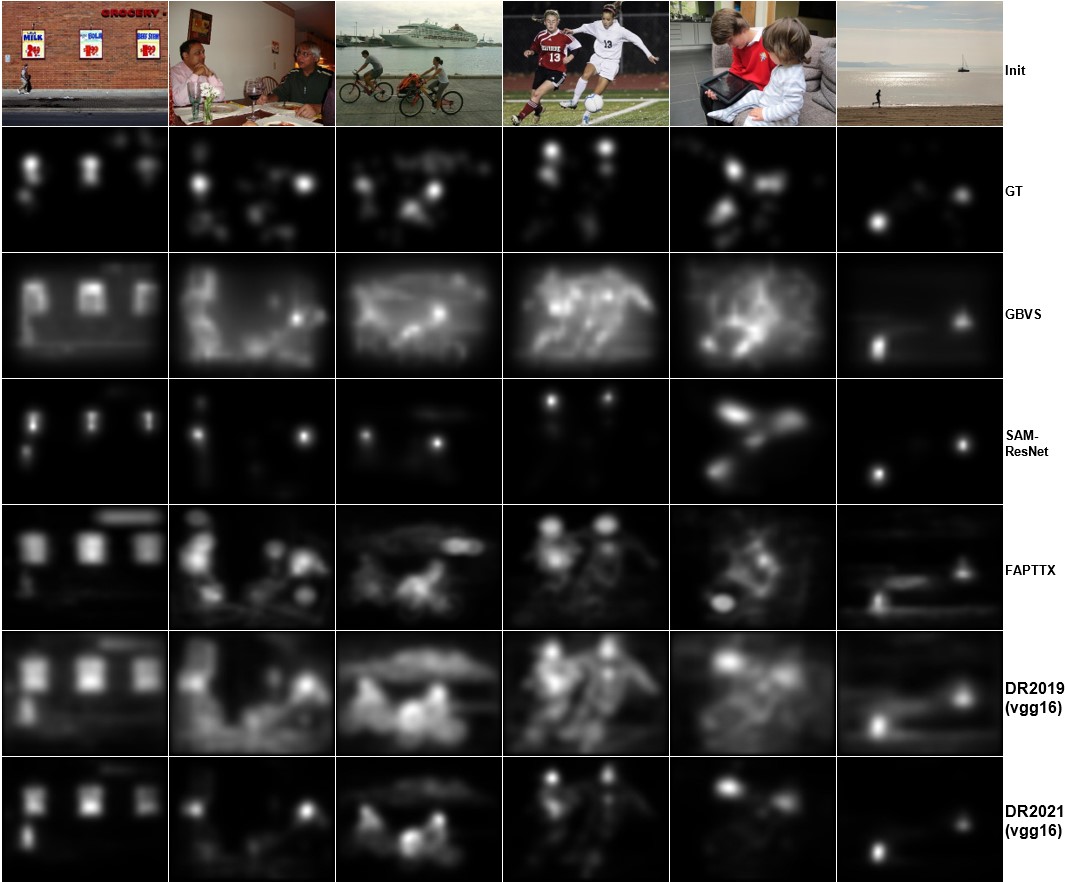}
\caption{Selected samples OSIE dataset. From top to down : initial, ground truth, GBVS, SAM-ResNet, FAPTTX \cite{kong19}, DR2019, DR2021.}
\label{fig:figOSIE}
\end{figure}

Figure \ref{fig:figOSIE} shows the images from OSIE dataset. \textbf{DR19} again spots the main correct salient regions but exhibits a lot of noise or distractors around them with a saliency map less focused as the one of the ground truth (line two). This issue is partially solved by \textbf{DR21} which is much more selective but still less than some DNN-based models. 

Overall, the qualitative study reveals that \textbf{DR21} spots most of the time the most important regions in all datasets. On MIT1003 and OSIE dataset the results of \textbf{DR21} are in most cases correct. If some DNN-based models are probably better on MIT1003 or OSIE datasets, one reason is that they are more focused on the top-down areas only as the ground-truth is. Indeed, DNN-based models were trained on images close the the ones in those two datasets. On O\textsuperscript{3} and P\textsuperscript{3} datasets, \textbf{DR21} clearly show their superiority on DNN-based models which are sometimes completely lost with very bad results. \textbf{DR19} and especially \textbf{DR21} exhibits the most stable behaviour performing well on all datasets while other models might be good on some images but much less good on others. 

\subsection{Quantitative validation on the different datasets}
We make a quantitative validation of different models based on the DeepRare framework on the four datasets shown in the previous section. 

First on MIT1003 and OSIE datasets which show general-purpose images where learning objects is very important. Those datasets should definitely provide an advantage to DNN-based models which focus on top-down information such as objects (faces, text, etc.) instead of bottom-up salient information. We previously showed in \cite{kong19} that the DNN-based models mainly learn which objects are most of the time attended which leads to good results on images implying a high amount of top-down information while they are very bad in purely bottom-up information. 

On the other side, we use P\textsuperscript{3} dataset from \cite{Kotseruba2019} which shows synthetic psycho-physical images with pop-out bottom-up objects which should work better for classical saliency models and even more with \textbf{DR19} and \textbf{DR21} models. 

Finally, we use O\textsuperscript{3} dataset from \cite{Kotseruba2019} which also provides real life images but with odd-out-one regions. The dataset is somewhere in the middle between P\textsuperscript{3} on one side and MIT1003 and OSIE datasets on the other side. O\textsuperscript{3} dataset should provide similar difficulty to classical and DNN-based saliency models.

\subsubsection{MIT1003 dataset evaluation}
\label{sec:mit}
We summarize in Table \ref{tab:omit} the results of \textbf{DR19} and \textbf{DR21} and also results coming from \cite{Kotseruba2019} for MLNet and SALICON models where MLNet was trained with SALICON, P\textsuperscript{3} and O\textsuperscript{3} datasets and SALICON was trained with OSIE, P\textsuperscript{3} and O\textsuperscript{3} datasets. The idea is to avoid trainings of MLNet or SALICON models on the MIT1003 dataset where it is evaluated to be fair towards unsupervised models and of course this gives lower results than the same models trained with the MIT1003 dataset. For other models (DeepFeat, eDN, GBVS, RARE2012, BMS, AWS), the figures come from \cite{deepFeat}. 

For \textbf{DeepRare} the following variants are used : \textbf{DR19-V16-WF} (\textbf{DR19} with a VGG16 backbone and without using the faces layer), \textbf{DR19-V16} (\textbf{DR19} with a VGG16 backbone and by using the faces layer), \textbf{DR21-MN2} (\textbf{DR21} with a MobileNetV2 backbone and without using faces information), \textbf{DR21-V16-WF} (\textbf{DR21} with a VGG16 backbone and without using the faces layer), \textbf{DR21-V16} (\textbf{DR21} with a VGG16 backbone and by using the faces layer), \textbf{DR21-V19} (\textbf{DR21} with a VGG19 backbone and without using faces information).  

The best model is definitely \textbf{DR21-V19} on all the metrics which is better than classical models but also than deep-features models (DeepFeat) and also all the DNN-based models in the Table \ref{tab:omit}. However, SALICON and MLNet were trained on datasets which are different from the MIT1003 training set which makes their performances lower than if they were trained on images from MIT1003.  

\begin{table}[!t]
\begin{center}
\caption{MIT1003 dataset. DeepRare2021 (VGG19 : DR21-V19, VGG16 without faces : DR21-V16-WF, VGG16 with faces : DR21-V16, MobileNetV2 : DR21-MN2), DeepRare2019 (VGG16 : DR19-V16), DeepRare2019 (VGG16 without faces : DR19-V16-WF, VGG16 with faces : DR19-V16), DFeat, eDN, GBVS, RARE2012, BMS, AWS results come from \cite{deepFeat} and SALICON and MLNet come from \cite{Kotseruba2019}.}
\label{tab:omit}
\begin{tabular}{|c|c|c|c|c|c|c|}
\hline 
Models & AUCJ $\uparrow$ & AUCB $\uparrow$ & CC $\uparrow$ & KL $\downarrow$ & NSS $\uparrow$ & SIM $\uparrow$ \tabularnewline
\hline 
\textbf{DR21-V19} & \textbf{0.86} & \textbf{0.85} & \textbf{0.56} & \textbf{0.88} & \textbf{1.93} & \textbf{0.50} \tabularnewline 
\hline 
DR21-V16 & 0.84 & 0.83 & 0.50 & 1.19 & 1.81 & 0.43 \tabularnewline
\hline 
DR21-V16-WF & 0.84 & 0.83 & 0.49 & 1.16 & 1.75 & 0.42 \tabularnewline
\hline 
DR21-MN2 & 0.84 & 0.83 & 0.50 & 1.14 & 1.71 & 0.42 \tabularnewline
\hline 
DR19-V16 & 0.86 & 0.85 & 0.48 & 1.25 & 1.58 & 0.36 \tabularnewline
\hline 
DR19-V16-WF & 0.84 & 0.83 & 0.46 & 1.32 & 1.54 & 0.34 \tabularnewline
\hline 
SALICON & 0.83 & - & 0.51 & 1.12 & 1.84 & 0.41\tabularnewline
\hline 
MLNet & 0.82 & - & 0.46 & 1.36 & 1.64 & 0.35\tabularnewline
\hline
DFeat & 0.86 & 0.83 & 0.44 & 1.41 & - & -\tabularnewline
\hline 
eDN & 0.86 & 0.84 & 0.41 & 1.54 & - & -\tabularnewline
\hline
GBVS & 0.83 & 0.81 & 0.42 & 1.3 & - & -\tabularnewline
\hline 
RARE2012 & 0.75 & 0.77 & 0.38 & 1.41 & - & -\tabularnewline
\hline 
BMS & 0.75 & 0.77 & 0.36 & 1.45 & - & -\tabularnewline
\hline 
AWS & 0.71 & 0.74 & 0.32 & 1.54 & - & -\tabularnewline
\hline 
\end{tabular}
\end{center}
\end{table}

\subsubsection{OSIE dataset evaluation}
We summarize in Table \ref{tab:osie} the results on the OSIE dataset. Here we added SAM-ResNet and FAPTTX models with the results reported in \cite{kong19}. SALICON and MLNet models are trained as in section \ref{sec:mit}. SAM-Resnet is used with its default training parameters showing that when trained on general images without introducing datasets such as P\textsuperscript{3} which can disturb the learning in general images cases, modern DNN-based models are better than DeepRare models in any version. FAPTTX also exhibits slightly better results showing the importance of top-down features in general images datasets. Our hypothesis here is that DeepRare models achieve better bottom-up scores than RARE2012 (verified on all datasets) but that the top-down information added to RARE2012 in FAPTTX makes it better. To verify this, we also added to DeepRare2021 using VGG16 the same top-down information (TD) than the one which was added to RARE2012 in \cite{kong19} and called this model DR21-V16+TD. This model is indeed better than FAPTTX proving that the top-down information is still missing from the DeepRare models. 

The same idea is once again illustrated by the fact that \textbf{DR19-V16} is better (on both MIT1003 and OSIE) than \textbf{DR19-V16-WF} even if the faces layer in VGG16 is much less efficient than a face detector as those used for FAPTTX. This again shows that DeepRare models do not capture top-down information which let room for future improvements.  

Another interesting point is that VGG16 backbone is slightly better for the OSIE dataset while VGG19 was better for MIT1003 showing that in MIT1003 maybe higher-level features are more important than in OSIE. A second point is about the fact that FAPPTX shows good results on this kind of images. FAPPTX is built upon RARE2012 with additional top-down features showing that adding top-down features to \textbf{DR21} would probably lead to results close to SAM-ResNet as \textbf{DR21} is better than RARE2012 in all configurations.   

\begin{table}[!t]
\begin{center}
\caption{OSIE dataset. DeepRare2021 (VGG19 : DR21-V19, VGG16 without faces : DR21-V16-WF, VGG16 with faces : DR21-V16, MobileNetV2 : DR21-MN2), DeepRare2019 (VGG16 : DR19-V16), DeepRare2019 (VGG16 without faces : DR19-V16-WF, VGG16 with faces : DR19-V16), and SAM-ResNet, FAPTTX, RARE2012, AWS, GBVS, and AIM come from \cite{kong19}. We added DeepRare2021 with VGG16 and top-down from \cite{kong19} called DR21-V16+TD.}
\label{tab:osie}
\begin{tabular}{|c|c|c|c|c|c|c|}
\hline 
Models & AUCJ $\uparrow$ & AUCB $\uparrow$ & CC $\uparrow$ & KL $\downarrow$ & NSS $\uparrow$ & SIM $\uparrow$ \tabularnewline
\hline
\textbf{SAM-ResNet} & \textbf{0.90} & \textbf{-} & \textbf{0.77} & 1.37 & \textbf{3.1} & \textbf{0.65}\tabularnewline
\hline
DR21-V16+TD & 0.88 & 0.83 & 0.66 & 0.83 & 2.32 & 0.56 \tabularnewline
\hline
FAPTTX & 0.87 & - & 0.62 & \textbf{0.81} & 2.08 & 0.51\tabularnewline
\hline
DR21-V16 & 0.87 & 0.86 & 0.59 & 0.91 & 2.06 & 0.52 \tabularnewline
\hline 
DR21-V16-WF & 0.87 & 0.86 & 0.58 & 0.84 & 2.01 & 0.51 \tabularnewline
\hline 
DR19-V16 & 0.87 & 0.86 & 0.55 & 0.98 & 1.75 & 0.44 \tabularnewline
\hline 
DR19-V16-WF & 0.86 & 0.86 & 0.53 & 1.01 & 1.66 & 0.43\tabularnewline
\hline 
DR21-MN2 & 0.85 & 0.84 & 0.51 & 1.06 & 1.55 & 0.42\tabularnewline
\hline
DR21-V19 & 0.83 & 0.82 & 0.45 & 1.32 & 1.54 & 0.34\tabularnewline
\hline 
RARE2012 & 0.83 & - & 0.46 & 1.05 & 1.53 & 0.43\tabularnewline
\hline 
AWS & 0.82 & - & 0.45 & 1.11 & 2.02 & 0.42\tabularnewline
\hline 
GBVS & 0.81 & - & 0.43 & 1.08 & 1.34 & 0.42\tabularnewline
\hline 
AIM & 0.77 & - & 0.32 & 1.52 & 1.07 & 0.34\tabularnewline
\hline 
\end{tabular}
\end{center}
\end{table}

\subsubsection{O\textsuperscript{3} dataset evaluation}
\label{subsec:o3d}
The O\textsuperscript{3} dataset uses the MSR metric defined in \cite{Kotseruba2019}. When the MSR\textsubscript{t} is higher, it is better as the target is well highlighted compared to the distractors. When MSR\textsubscript{b} is lower, it is better, it means that the maximum of the saliency of the target is higher than the one of the background. The first measure will ensure that the target is visible compared to the distractors and the second that it is visible compared to the background.  

Table \ref{tab:op3} shows the MSR measures from the paper of \cite{Kotseruba2019} where we added the results from the DeepRare models (\textbf{DR19} and \textbf{DR21} in the versions using VGG16, VGG19 and MobileNetV2 architectures) splitting the dataset between the images where color is a good discriminator (Color) and the others (Non-color). All models work better for targets where color is an important feature and less well for non-color. 

\begin{table}[!ht]
\small
\begin{center}
\caption{Comparing result between several models (SAM-Resnet, CVS, DeepGaze II, FES, ICF and BMS) and DR family (\textbf{DR19} and \textbf{DR21} in the version VGG16, VGG19 and MobileNetV2). For MSR\textsubscript{t} higher is better, For MSR\textsubscript{b} lower is better.} 
\label{tab:op3}
\begin{tabular}{|c|c|c|c|c|c|c|}
\hline 
{Models} & \multicolumn{2}{c|}{Color} & \multicolumn{2}{c|}{Non-color} & \multicolumn{2}{c|}{All targets}\tabularnewline
\cline{2-7} 
 & MSR\textsubscript{t} $\uparrow$ & MSR\textsubscript{b} $\downarrow$ & MSR\textsubscript{t} $\uparrow$ & MSR\textsubscript{b} $\downarrow$ & MSR\textsubscript{t} $\uparrow$ & MSR\textsubscript{b} $\downarrow$ \tabularnewline
\hline 
\textbf{DR21-V16} & \textbf{1.66} & \textbf{0.74} & \textbf{1.31} & 1.31 & \textbf{1.45} & \textbf{1.01}\tabularnewline 
\hline 
DR21-V19 & 1.63 & 0.78 & 1.29 & 1.39 & 1.43 & 1.13 \tabularnewline
\hline 
DR21-MN2 & 1.19 & 1.02 & 1.06 & 1.54 & 1.12 & 1.32 \tabularnewline
\hline 
DR19 & 1.14 & 0.75 & 1.00 & \textbf{1.00} & 1.06 & 0.89 \tabularnewline
\hline 
SAM-ResNet & 1.47 & 1.46 & 1.04 & 1.84 & 1.40 & 1.52\tabularnewline
\hline 
CVS & 1.43 & 2.43 & 0.91 & 4.26 & 1.34 & 2.72\tabularnewline
\hline 
DGII & 1.32 & 1.55 & 0.94 & 1.95 & 1.26 & 1.62\tabularnewline
\hline 
FES & 1.34 & 2.53 & 0.81 & 5.93 & 1.26 & 3.08\tabularnewline
\hline 
ICF & 1.30 & 2.00 & 0.84 & 2.03 & 1.23 & 2.01\tabularnewline
\hline 
BMS & 1.29 & 0.97 & 0.87 & 1.59 & 1.22 & 1.07\tabularnewline
\hline 
\end{tabular}
\end{center}
\end{table}

For MSR\textsubscript{t} (higher is better) for Color \textbf{DR19} is less good especially compared to DNN-based models. However we can see that for Non-color images where the models fail much more \textbf{DR19} has a remarkable stability being second and very close the the best one (SAM-ResNet). \textbf{DR21} especially using the VGG19 and VGG16 architectures are definitely the best models being much better even than efficient DNN-based modes such as SAM-ResNet on all the kinds of images. 

If we take into account the MSR\textsubscript{b} (lower is better), the DeepRare models clearly outperforms all the others providing the best discrimination between the target and the background. DeepRare models are the only ones with a MSR\textsubscript{b} smaller than 1 which means that in average the maximum of the target saliency is higher than the maximum of the background saliency. \textbf{DR21} with VGG16 architecture is still better than all classical and DNN-based models and even better than \textbf{DR19} for Color images. 

In conclusion, for MSR\textsubscript{t} and MSR\textsubscript{b} metrics, the models from the DeepRare family and especially \textbf{DR21} with VGG16 architecture outperform all the other models including efficient DNN-based models on both Color or Non-color images on O\textsuperscript{3} dataset.   

\begin{table}[!ht]
\begin{center}
\caption{SALICON, MLNet and DeepRare family (\textbf{DR19} and \textbf{DR21} with MobileNetV2, VGG19 and VGG16 architectures) results on O\textsuperscript{3} dataset.}
\label{tab:op3o3}
\begin{tabular}{|c|c|c|}
\hline 
Models & MSR\textsubscript{t} $\uparrow$ & MSR\textsubscript{b} $\downarrow$\tabularnewline
\hline 
\textbf{DR21-V16} & \textbf{1.45} & \textbf{1.01}\tabularnewline
\hline 
DR21-V19 & 1.43 & 1.13\tabularnewline
\hline 
DR21-MN2 & 1.12 & 1.32\tabularnewline
\hline 
DR19 & 1.06 & 0.89\tabularnewline
\hline 
MLNet & 0.96 & 0.91\tabularnewline
\hline 
SALICON & 0.90 & 1.26\tabularnewline
\hline 
\end{tabular}
\end{center}
\end{table}

Table \ref{tab:op3o3} shows the results of the DeepRare family compared to two other DNN-based models tested on the whole O\textsuperscript{3} dataset (both Color and Non-color images). Our models outperform both SALICON and MLNet models on both MSR\textsubscript{t} (all the DeepRare models are better) and MSR\textsubscript{b} (\textbf{DR19} is better) metrics. According to \cite{Kotseruba2019}, the results we show here for SALICON are the ones where it was trained on the OSIE by adding with P\textsuperscript{3} and O\textsuperscript{3} datasets. The MLNet was trained on SALICON by adding with P\textsuperscript{3} and O\textsuperscript{3} datasets.

\subsubsection{P\textsuperscript{3} dataset evaluation}
The P\textsuperscript{3} dataset is the one which exhibits the less top-down information and it even does not have any centered bias. Naturally, for this dataset, the DNN-based models perform the worst. We will check here how the DeepRare models deal with the data. 

First we use the average \# of fixations and found percentage metrics. The average \# of fixations is better if lower as it means that the target is found more rapidly and the found percentage metric is better if higher as it means that a higher percentage of the target is found after 100 fixations. Table \ref{tab:op4} shows first the results on P\textsuperscript{3} for DeepRare models compared with SALICON and MLNet models. For the SALICON and MLNet models they were trained the same way than in section \ref{subsec:o3d}. Our models all definitely outperform the two DNN-based models and need much less fixations to discover more of the targets showing here very good results.  

\begin{table}[!ht]
\begin{center}
\caption{Comparing result on P\textsuperscript{3} dataset.}
\label{tab:op4}
\begin{tabular}{|c|c|c|}
\hline 
Model & Avg. \# fix. $\downarrow$ & \% found $\uparrow$ \tabularnewline
\hline 
\textbf{DR21-V16} & \textbf{13.53} & \textbf{89} \tabularnewline
\hline 
DR21-V19 & 13.86 & 89 \tabularnewline
\hline 
DR21-MN2 & 33.82 & 72 \tabularnewline
\hline 
DR19 & 16.34 & 87 \tabularnewline
\hline 
MLNet & 42.00 & 44\tabularnewline
\hline 
SALICON & 49.37 & 65\tabularnewline
\hline 
\end{tabular}
\end{center}
\end{table}

Table \ref{tab:op6} provides more details about the found percentage metric after different numbers of fixations (15, 25, 50 and 100) and for specific images where the target is due to color, orientation or size features with 100 fixations. The results here are compared with classical models which are better in this dataset than DNN-based models. On this table DeepRare models are the best again and especially \textbf{DR21} with the VGG16 architecture. While BMS can exhibit 100\% for color or orientation target percentage found, it is more efficient in terms of detection to find the target (even if not its entire surface) very quickly (15 fixations) than to find all of the target surface but after 100 fixations. So if we look at the results after 15 fixations only the DeepRare methods are all much better than the others.       

\begin{table}[!t]
\begin{center}
\caption{Comparing result on P\textsuperscript{3} dataset. Details on the percentage found after the number of fixation of 15 (\%fd15), 25 (\%fd25), 50 (\%fd50), and 100 (\%fd100). Percentage found of the color ((\%fd-C), orientation (\%fd-O), and size (\%fd-S) features taken separately.}
\label{tab:op6}
\begin{tabular}{|c|c|c|c|c|c|c|c|}
\hline 
Model & \%fd15 & \%fd25 & \%fd50 & \%fd100 & \%fd-C & \%fd-O & \%fd-S\tabularnewline
\hline 
\textbf{DR21-V16} & \textbf{84.82} & \textbf{86.71} & \textbf{88.60} & \textbf{89.76} & \textbf{92.20} & \textbf{92.93} & \textbf{83.92}\tabularnewline
\hline 
DR21-V19 & 84.27 & 86.32 & 88.10 & 89.14 & 92.65 & 92.36 & 82.14\tabularnewline
\hline 
DR21-MN2 & 61.37 & 64.81 & 69.37 & 72.46 & 77.17 & 71.75 & 68.21\tabularnewline
\hline 
DR19 & 80.61 & 83.27 & 86.63 & 87.87 & 91.29 & 89.58 & 82.50\tabularnewline
\hline 
RARE2012 & 59.87 & 63.52 & 79.75 & 93.48 & 99.54 & 90.26 & 88.53\tabularnewline
\hline 
BMS & 58.94 & 66.37 & 83.56 & 95.14 & \textbf{100} & \textbf{100} & 82.76\tabularnewline
\hline 
ICF & 32.63 & 41.38 & 68.47 & 70.18 & 69.41 & \textbf{100} & 42.45\tabularnewline
\hline 
oSALICON & 30.25 & 39.75 & 55.45 & 78.53 & 76.35 & 81.58 & 70.42\tabularnewline
\hline 
\end{tabular}
\end{center}
\end{table}

\begin{figure}[!ht]
\centering
\includegraphics[width=4in]{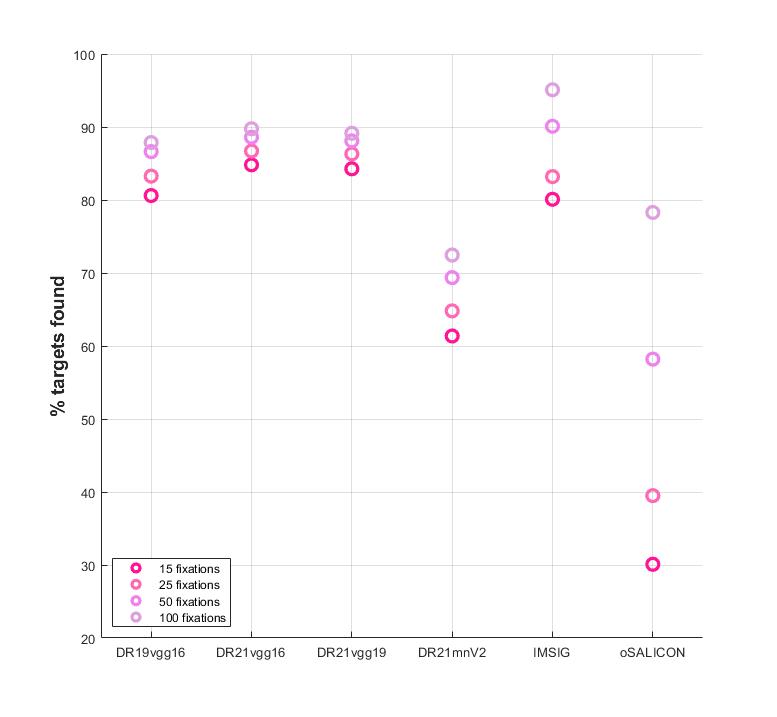}
\caption{Number of fixations (horizontal axis) vs. \% of targets detected (vertical axis). It is chosen on 15, 25, 50, and 100 fixations.}
\label{fig:nb100}
\end{figure}

Figure \ref{fig:nb100} shows the DeepRare family models compared to the best classical model (IMSIG) and the best DNN-based model (oSALICON). If we look at the percentage of targets found after only 15 fixations, than \textbf{DR21} with the VGG16 and VGG19 architectures are the best followed by \textbf{DR19}, IMSIG and oSALICON which is definitely worse. oSALICON (OpenSALICON), refers to \cite{Kotseruba2019} the saliency maps are obtained using the pre-trained OpenSALICON weights on the SALICON dataset. In that way oSALICON is not trained on the P\textsuperscript{3} dataset again to remain fair.    

Finally, the GSI metric (Global Saliency Index) is computed on this dataset. This score is better when higher as it measures how target average saliency is distinguished from the distractors. For GSI, table \ref{tab:op5} shows the average figures for the whole dataset (GSI-Avg) and for each of the dataset classes : images where the feature of the target is based on color (GSI-Color), on the orientation (GSI-Orientation) and on size (GSI-Size). The average scores for the GSI metric are much higher for the DeepRare models and especially for \textbf{DR21} with the VGG16 architecture. While the results of classical models such as BMS or RARE2012 can be comparable on GSI-Color, for GSI-Orientation or GSI-Size they are much less good than those of the DeepRare models. If we take into account the DNN-based models, than the GSI scores begin to be even negative, showing that distractors are in average more visible than the salient areas indicating that DNN-based models do not work at all here.  

\begin{table}[!t]
\begin{center}
\caption{Comparing result on P\textsuperscript{3} dataset. Global Saliency Index score on color, orientation, and size features, and average score from these 3 features.}
\label{tab:op5}
%\begin{adjustbox}{width=1\textwidth}
%\small
\begin{tabular}{|c|c|c|c|c|}
\hline 
Model & GSI-Color & GSI-Orientation & GSI-Size & GSI-Avg.\tabularnewline
\hline 
\textbf{DR21-V16} & \textbf{0.77} & \textbf{0.50} & 0.49 & \textbf{0.59} \tabularnewline
\hline 
DR21-V19 & 0.75 & 0.49 & \textbf{0.51} & 0.58\tabularnewline
\hline 
DR21-MN2 & 0.66 & 0.42 & \textbf{0.51} & 0.53\tabularnewline
\hline 
DR19 & 0.42 & 0.17 & 0.15 & 0.25\tabularnewline
\hline 
RARE2012 & 0.74 & 0.01 & 0.18 & 0.31\tabularnewline
\hline 
BMS & 0.72 & 0.01 & -0.02 & 0.24\tabularnewline
\hline 
ICF & 0.18 & -0.02 & -0.51 & -0.12\tabularnewline
\hline 
oSALICON & -0.01 & 0.04 & -0.11 & -0.03\tabularnewline
\hline 
\end{tabular}
%\end{adjustbox}
\end{center}
\end{table}

\begin{figure}[!ht]
\centering
\includegraphics[width=5in]{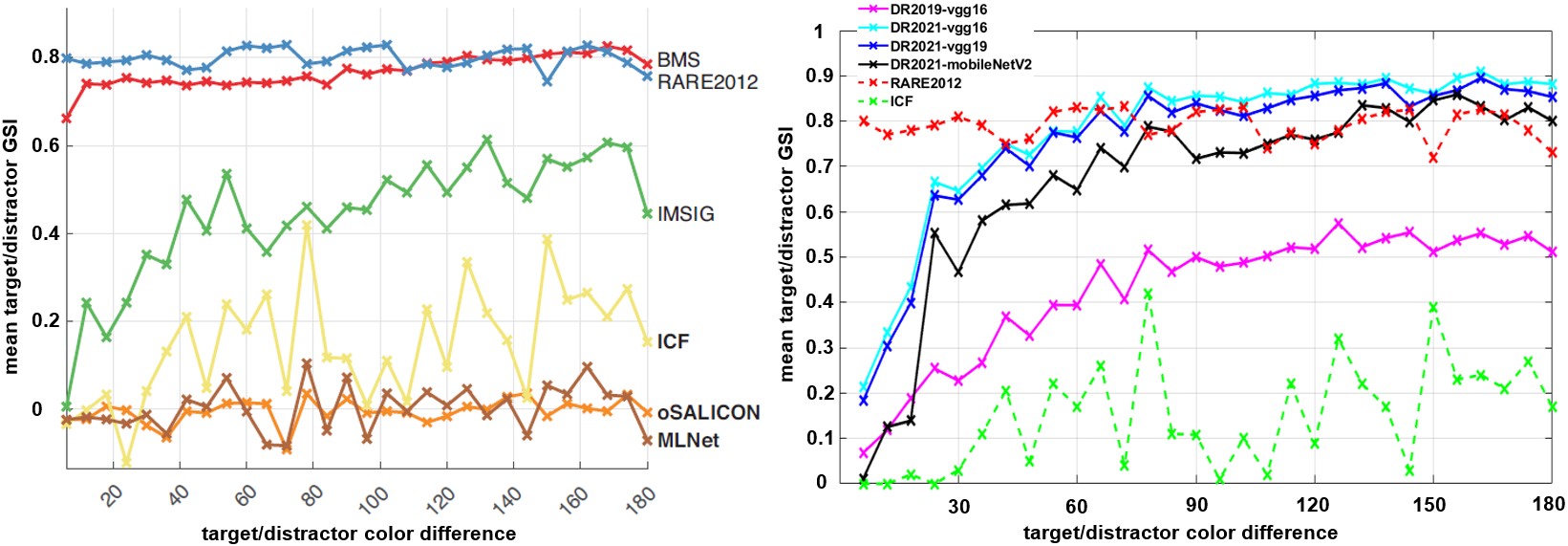}
\caption{The GSI score for color target/distractor difference. Left plot: generated by \cite{Kotseruba2019}. Right plot: several classical and deep learning models including our DR2021 model.}
\label{fig:gsic}
\end{figure}

\begin{figure}[!ht]
\centering
\includegraphics[width=5in]{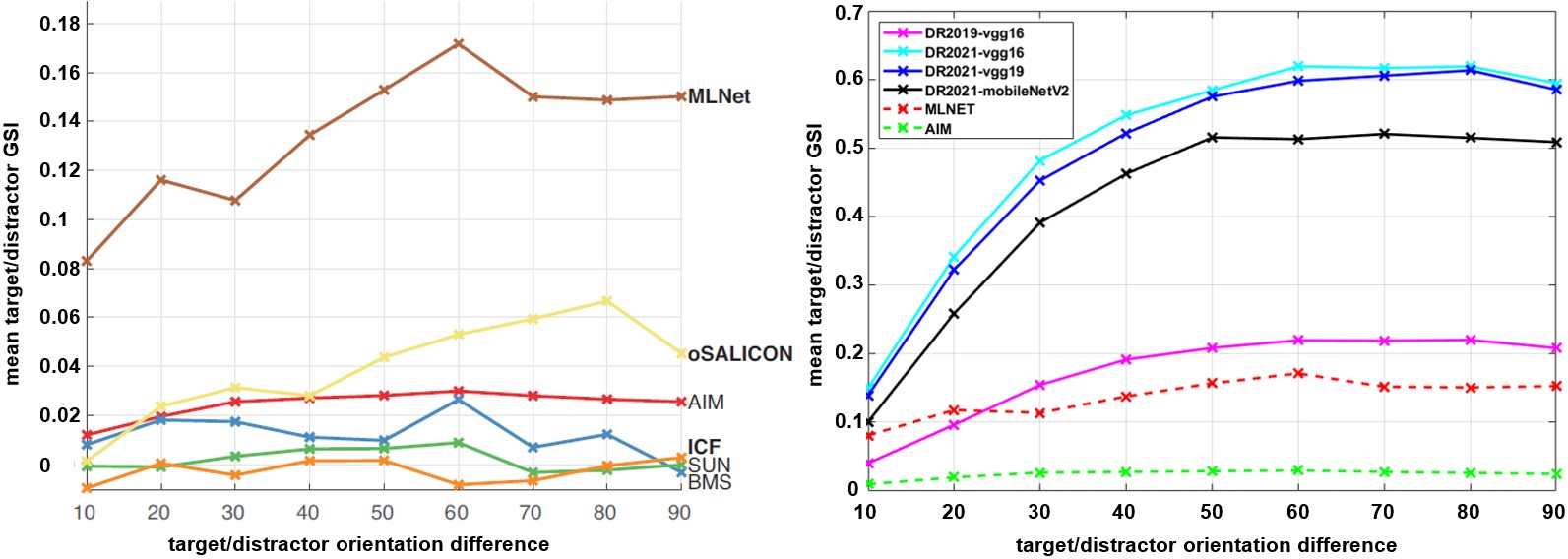}
\caption{The GSI score for orientation target/distractor difference. Left plot: generated by \cite{Kotseruba2019}. Right plot: several classical and deep learning models including our DR2021 model.}
\label{fig:gsio}
\end{figure}

\begin{figure}[!ht]
\centering
\includegraphics[width=5in]{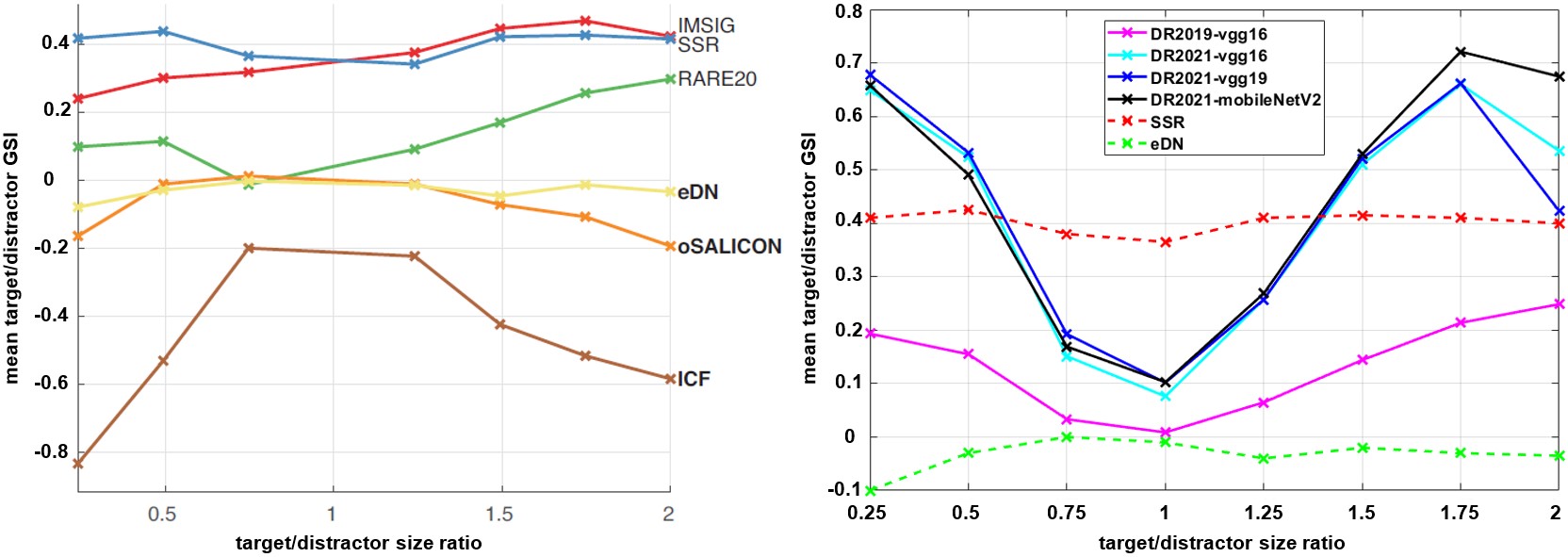}
\caption{The GSI score for size target/distractor size ratio. Left plot: generated by \cite{Kotseruba2019}. Right plot: several classical and deep learning models including our DR2021 model.}
\label{fig:gsis}
\end{figure}

Figures \ref{fig:gsic}, \ref{fig:gsio} and \ref{fig:gsis} let us compare the dynamics of the GSI scores on the three classes of models (GSI-Color, GSI-Orientation and GSI-Size). For each figure, we show the three best classical models with the three best DNN-based models (name in bold) on the left and DeepRare models results with the best classical and the best DNN-based model on the right. 

For color targets (figure \ref{fig:gsic}, right graph) we see that the maximum of GSI score for \textbf{DR21} with a VGG16 architecture where GSI is at more than 0.9. If RARE2012 model is better on small target/distractor color difference, \textbf{DR21} is better for larger differences. The ICF model is less good than all the other models from the DeepRare family on any target/distractor color difference. We also see that \textbf{DR21} model is better than \textbf{DR19} for all used architectures. 

In addition, the shape of the GSI curve exhibited by the DeepRare family of models is coherent from a biological point of view: if the difference between the target color and the distractor color is small, then the model detects less well the target (left-side of the curve) than when the color of the target and background is very different (right-side of the curve). The models from the DeepRare family are the only ones to provide a biologically plausible GSI curve. 

For orientation targets (figure \ref{fig:gsio}, right graph) we see that the maximum of GSI score for \textbf{DR21} with a VGG16 architecture is at more than 0.6 (right graph). This score is drastically higher than the best DNN-based model and the best classical model on all target/distractor orientation difference. We also remark again that \textbf{DR21} model is better than \textbf{DR19} for all used architectures.    

Also, the shape of the GSI curve exhibited by DeepRare family models is again coherent from a biological point of view: if the difference between the target orientation and the distractor orientation is small (left-side of the curve), then the model detects the target less well than when target orientation is very different from the distractors (right-side of the curve). Here also only the DeepRare family models have a dynamic which is close to the one expected from a human.  

For size targets (figure \ref{fig:gsis}, right graph) we see that the maximum of GSI score for the best model (\textbf{DR21} with a MobileNetV2 architecture) is about 0.7 which makes it close to RARE2012 in terms of maximum GSI. The best classical model (SSR) is less good when the target/distractor size ratio is smaller or bigger (left-side or right-side of the curve) but better when this ratio is close to 1 where there is a small difference between the target and the distractors (center of the graph). The best (here eDN) is much worse than the DeepRare family models on any target/distractor ratio. \textbf{DR21} with any architecture is again much better here than \textbf{DR19}.

The shape of the GSI curve exhibited by our model is finally again coherent from a biological point of view: if the difference between the target size and the distractor size is small (center of the curve), then the model detects the target less well than when its size is very different (left and right sides of the curve). We can also see an asymmetry in the curve showing that it is easier for \textbf{DR19} to detect target twice bigger than distractors than targets twice smaller than the distractors which is again biologically coherent. This is also true for \textbf{DR21} even if for very big target size (2 times bigger than the distractors) we can see a decrease in the performance.  

\section{Discussion and Conclusion}
\label{sec:disc}
We proposed a novel saliency fraemwork called DeepRare using the simplified rarity idea of \cite{rare2012} applied on the deep features extracted by a deep neural network pre-trained on ImageNet dataset. After a first instantiation of this framework called DeepRare2019 we propose here DeepRare2021 which exhibits several interesting features: 
\begin{itemize}
\item It needs no training, and the default ImageNet training is enough. ImageNet is a  generic image dataset which let DNN encoders extract most of the useful image-related features needed to understand images (ImageNet is trained for objects classification). 
\item DeepRare2021 introduces several novelties compared to DeepRare2019 among which the use of thresholded rarity maps which drastically improve the results in terms of performance compared to DeepRare2019. The use of the thresholded maps makes DeepRare2021 much less sensitive to distractors allowing it to focus more on the main surprising areas. 
\item The model is computationally efficient and is easy to run even on CPU only. 
\item In comparison with DeepRare2019 where only a VGG16 architecture could fit to the model, the DeepRare2021 approach is very modular, and it is easy to adapt to any neural network architecture such as VGG16, VGG19, or even more complex architectures such as MobileNetV2 for adaptation on mobile devices as smartphones or for edge computing.
\item It is possible to check each layer contribution to the final saliency map and thus better understand the result. It is also possible to check several thresholds to see which areas of the images are considered as the most rare compared to the others and at which levels. This opportunity is a key feature of DeepRare2021 contrary to DeepRare2019 and even more contrary to black-box DNN-based models. Indeed, if DeepRare2021 does not work well, it is easy to segment the different layer maps and understand from which layer the issue comes from, or if the issue comes from the fusion step. In a specific case, for example, the model was not able to find a surprising object which was very big into the image. While looking to the decomposition of the different layers such the ones that can be seen in Figure \ref{fig:feat3}, only one level (the higher) detected the surprising object, but the final saliency map was not highlighting it because all the other levels were not detecting this object, so the fusion step was the issue in this specific case.     
\item DeepRare models are very generic and stable through all kinds of different datasets where other models are sometimes better but only for one dataset and/or a specific metric but much worse for the others. The DeepRare2021 version is specifically better than DeepRare2019 on all datasets when compared with the same VGG16 architecture. DeepRare2021 is thus the most generic model and when applying it to a new and unknown dataset it will surely provide results which make sense while with DNN-based models, there is no certitude that on a new image dataset it will provide meaningful results (especially if the dataset is not close to the ones used for training). If this is not a crucial issue on natural images which are more or less close to the training datasets, for specific datasets such as images with defects for industrial quality control DNN-based models will perform very poorly. In addition, if the defects do not have a specific shape, even by re-training the DNN-based models, they will not be able to learn defects with various shapes as rare features are very hard to learn by definition. If defects attract human gaze, it is specifically because they are unknwon and surprising and humans are not able to learn them if they do not repeat in the same way.        
\end{itemize}

We show that this framework, especially DeepRare2021, is the most stable and generic when testing it on 4 very different datasets. It was first tested on MIT1003 and OSIE where it outperforms all the classical models and most of the DNN-based models. However some DNN-based models, especially the latest ones, still provide better results. 

We then tested DeepRare models on the O\textsuperscript{3} dataset, where DeepRare2021 outperforms all the models on target/background discrimination and on target/distractor discrimination. Finally, on P\textsuperscript{3} dataset, our model is first for the target discrimination based on the number of fixations. When computing the average GSI metric our model is also the best for all the features (color, orientation, size) and the only one to exhibit a GSI plot which is biologically plausible.

While one cannot expect from an unsupervised model such as DeepRare models to be better on MIT1003 or OSIE dataset than DNN-based models which are trained and tuned on similar data, those DNN-based models are bad or even completely lost on O\textsuperscript{3} and P\textsuperscript{3} datasets and on any dataset containing surprising areas which have various shapes and thus cannot be learnt by DNN architectures.  

Our tests show that DeepRare models and especially DeepRare2021 models are optimized models overcoming any classical model and being only beaten by recent DNN-based models on MIT1003 or OSIE datasets. They are generic, unsupervised and stable in results on all kind of datasets. Even if they take into account low- and high-level features, they still remain bottom-up approaches as FAPTTX results show \cite{kong19}. Indeed by adding top-down information to RARE2012, the results of FAPTTX are still comparable or a little better than for the DeepRare models. However if we add the same top-down information as in \cite{kong19} to DeepRare21 instead of Rare2012, DeepRare with top-down outperforms Rare2012 with top-down information. The fact that top-down information is important can also be seen with the fact that DR21-V16 is most of the time better than DR21-V16-WF because it uses information about faces.  

This remark leads to future works for future implementations of the DeepRare models. Adding top-down information on top of DeepRare2021 would probably drastically improve its performance on MIT1003 and OSIE datasets while keeping similar results on  O\textsuperscript{3} and on P\textsuperscript{3} datasets. 

The DeepRare family framework shows that deep-features-engineered models might become a good choice in visual attention field especially when the 1) images they are applied on are special and specific and 2) eye-tracking datasets are not available on this kind of images or when 3) explaining the result is of high importance for example the case of industrial standardization.  

% use section* for acknowledgement
\section*{Acknowledgments}
%Supported by anonymous funds.
Supported by ARES-CCD (program AI 2014-2019) under the funding of Belgian university cooperation.

\bibliography{mybibfile}

\end{document}